# Towards Highly Expressive Machine Learning Models of Non-Melanoma Skin Cancer

S.M.Thomas[a]   J.G.Lefevre[a]   G.Baxter[b]   N.A.Hamilton[a]

[a] Institute for Molecular Bioscience, The University of Queensland, St Lucia, Australia
[b] MyLab Pathology, Salisbury, Australia

# Abstract

Pathologists have a rich vocabulary with which they can describe all the nuances of cellular morphology. In their world, there is a natural pairing of images and words. Recent advances demonstrate that machine learning models can now be trained to learn high-quality image features and represent them as discrete units of information. This enables natural language, which is also discrete, to be jointly modelled alongside the imaging, resulting in a description of the contents of the imaging. Here we present experiments in applying discrete modelling techniques to the problem domain of non-melanoma skin cancer, specifically, histological images of Intraepidermal Carcinoma (IEC). Implementing a VQ-GAN model to reconstruct high-resolution (256×256) images of IEC images, we trained a sequence-to-sequence transformer to generate natural language descriptions using pathologist terminology. Combined with the idea of interactive concept vectors available by using continuous generative methods, we demonstrate an additional angle of interpretability. The result is a promising means of working towards highly expressive machine learning systems which are not only useful as predictive/classification tools, but also means to further our scientific understanding of disease.

# Introduction

## The Vocabulary of Pathology

Histological images contain a vast amount of information. Across low and high magnifications, diverse and contrastive perspectives of microscopic tissue structures can be observed. For the pathologist, such variable and complex information is best understood and expressed using natural language. Indeed, routine work involves jointly recognising histological patterns and communicating what they see using a large and highly specialised vocabulary. This forms the basis of diagnostic reports, which contain descriptions of the underlying cellular morphology/pathology, and act as a form of evidence gathering to inform a final diagnosis. There is thus a natural pairing of words and images in diagnostic pathology

The level of detail and understanding required sets a high bar for what constitutes knowledge in this domain. Although it is often perceived as a classification problem (malignant versus benign), the pathologist's ability to make explicit their knowledge underscores a robust diagnostic intuition, which is based on a highly adapted framework of scientific explanation. The same cannot be said about diagnostic machine learning systems which purportedly perform at the expert-human level [1, 2, 3, 4]. However, a system which could be similarly expressive would not only be more interpretable (and so allay concerns surrounding trust) but would be a step closer to what we really mean by *pathologist-level*.

This was similarly argued in 2017 with researchers developing an image classification and natural language generation model for characterising histological images of bladder cancers [5]. They combined CNNs for image processing and classification, with attention-based recurrent neural networks for text-generation. The image captions were composed of a limited set of descriptions, which were expressive enough to characterise the variety of relevant features. They subsequently went on to refine this method, culminating in a system which matches pathologist-level classification [6], and extended it to image captioning and retrieval problems in general [7]. Similar approaches have been followed for problems in radiology [8, 9].

Besides the work of Z. Zhang et al. [5] and derivatives, applications to histology are few. The main challenge is in curating and annotating datasets, due to the expertise needed in both the machine learning and medical domains. This is because relevant annotations are difficult to provide without a clear problem definition and envisioned application. This is discussed by Lindman et al. [10], who developed an annotated dataset of WSIs of colon (101) and skin tissue (99), following the SNOMED CT medical ontology [11]. The method involved labelling large tissue compartments and other features of interest in a systematic way using a controlled vocabulary derived from the ontology. However, the annotations were composed of terms and descriptions across various levels of magnification, which as mentioned may make them unsuitable for a specific task or problem. Ideally, descriptions should be tailored for a specific context as in the work of Z. Zhang et al. [5].

Another challenge is the cost of creating a dataset to the sizes required for deep learning applications to avoid learning spurious correlations. There is also the difficulty of having text-models reliably generate relevant and coherent sentences. Consequently, modelling a pathologist's ability to characterise image features in natural language remains an open problem.

## Recent Technological Advances

The recent release of the VQ-GAN [12], CLIP [13], and DALI [14] models, provided an impressive ability to analyse or generate images in combination with natural language text prompts. Specifically, VQ-GAN (where VQ stands for vector quantized) is a fully convolutional network which learns a discrete/quantized feature space, from which it tries to reconstruct the original image. An adversarial training regime enables rich features to be compressed into a code of comparatively small size e.g., 256×256 → 16×16.  CLIP (Contrastive Language–Image Pre-training) jointly learns image features and natural language captions using a contrastive learning approach using CNNs and large transformers [15]. The aim is to learn a similar latent representation for both the image and caption, so that related features and captions are mapped to similar locations. DALI is another discrete model (specifically a VQ-VAE), which compresses images down to a 32×32 representation. The discretized image representation and accompanying image caption are jointly fed to a large generative transformer which is trained to predict the next caption token, conditioned on the preceding sequence of both image and text tokens. The result is a generative model which can produce images that correspond to a text prompt.

Collectively these works have inspired a new wave of interest in combining text and images in a machine learning setting. However, the success of these models is underpinned by large datasets and large compute resources. The question of whether this can be achieved on smaller datasets and resources of more modest size is not known. Specifically, instead of large, generalized models trained on noisy and unstructured data, can meaningful success be found in a more constrained medical image context?

## The Continuous Morphological Progression of Skin Cancer

Previous work shows that histology images of skin cancer are amenable to machine learning methods, with the most commonly studied being basal cell carcinoma [1, 16, 17, 18]. The histological presentation usually falls into the binary categories of cancer and non-cancer, a consequence of the abnormal and obvious proliferation of basal cells. In contrast, the much less common Intraepidermal Carcinoma (IEC), is a diagnosis given upon observing a certain severity of dysplasia in the keratinocytes. After environmental exposure, normal keratinocytes may undergo morphological changes, which can be placed along a spectrum between normal, mild, moderate, and severe dysplasia [19]. These are considered precancerous morphological expressions and are collectively diagnosed as Actinic Keratosis (AK). It is only when the dysplastic cells proliferate completely from the basal to the granular layer of the epidermis, referred to as full-thickness dysplasia, is the lesion considered an IEC [19].

Understanding the degree of dysplasia in cancer is routine for pathologists, and has been modelled previously for prostate cancer using machine learning models to classify images into Gleason Grade categories [20, 21]. However, classification-only approaches leave much to be desired in terms of model interpretability. In contrast, the ability to explicitly visualise what the model has learned by generative modelling can help solve this problem [18], as well as provide predictive capabilities if desired. Ideally, using generative methods to model the continuous progression between normal epidermis to severe dysplasia, would not only identify the features representative of known categories, but could also aid in illuminating a new understanding of disease progression generally.

## Combining Discrete and Continuous Modelling

In this work, we attempt to build upon the success of recent developments of generative machine learning, combining the benefits of both discrete and continuous modelling methods. We aimed to develop a generative model to learn the continuous morphological relationships between tissue features in images of IEC. Central to this is the need for high-quality and high-resolution images, requiring us to create a new dataset of 256×256-pixel images. The ability to encode realistic images to and generate realistic images from a continuous representation, enables images to be manipulated and explored in meaningful ways via latent space interpolation and concept vectors. The other aspect of the problem is using natural language to characterise the features in the images. Therefore, we additionally create natural language summaries of images from our dataset using a controlled vocabulary of anatomical pathology terms. Jointly learning a discrete representation of the rich feature summaries enabled us to work towards the complimentary ability to produce natural language descriptions for the image content. This work therefore represents the ways this central part of pathology can be currently modelled with machine learning.

# Methods

## Image Dataset

*The data collected for this study was approved for use by the Office of Research Ethics at the University of Queensland, application number 2018001029.*

One aim of generative modelling is to learn a representation which summarises a hierarchy of features which can be used to produce realistic images. This is seen in progressively grown architectures [22, 23, 24] and their descendants [25, 26, 27], which show that models are capable of learning finer and finer details as image resolution increases. Applications to histological imaging show that generative models can successfully be trained at resolutions of 128×128 pixels [18], 224×224 pixels [28], 512×512 pixels [29], and 1024×1024 pixels [30], across a variety of cancer and tissue contexts. With the aim of learning the most relevant features, it is arguably desirable to model higher-resolution images. However, such goals can be limited by the practical challenges of training in terms of both stability and computational cost. Further, unlike images of human faces, bedrooms, or cars, which contain variations of a central object from limited perspectives, previous work in histological imaging covers a variety of tissue types across the full range of translations and rotations. This makes the problem more like texture synthesis in contrast to scene generation. Significantly, this impacts the degree to which we can interpret the model's latent space, where visualised journeys are unrealistic [18, 31]. In contrast, data sets with fixed perspectives provide parsimonious transitions between regions of the latent space, where intermediate images link meaningfully to the end points. Thus, to facilitate the synthesis of higher-resolution images as well as make latent space interpretability techniques viable, we created a new dataset that aimed to represent the image domain from a fixed and consistent perspective.

In early 2020, MyLab Pathology provided access to their pre-existing collection of 168 skin cancer slides representing typical cases of IEC and AK specimens. The set included shave (74), punch (26) and excision (68) biopsies, which were hand-annotated by a pathologist to indicate which tissue section was most representative of the diagnostic class. The respective tissue sections were manually imaged using a Leica DM microscope at 20× magnification with a C-Mount Basler aca2040-55uc camera of resolution 2048×1536 pixels. Each image was stitched into a whole slide image (WSI) in real-time using the manualWSI software by Microvisioneer (https://www.microvisioneer.com/). The resulting WSIs were saved as a hierarchical image in the .svs format (Aperio ScanScope slide scanner standard). At the lowest level (highest resolution) 1 pixel width corresponds to 0.314μm in length. The final image resolutions were in the range of 8K×8.5K to 42K×70K pixels.

As skin cancer emanates from the epidermis, a context window within each WSI was chosen which focused specifically on the epidermal layer with small portions of papillary dermis (dermal) and keratin on either side. In a sliding-window fashion, image patches of size 1024×1024 pixels were captured by following the epidermal layer along the length of the entire specimen with a step size of approximately 340 pixels (2/3 overlap). This path attempts to capture the continuous transformation from healthy epidermis to mild and severe dysplasia. At this resolution the morphological details such as cell junctions and mitotic events could be clearly distinguished. However, in consultation with a pathologist, it was decided that down-sampled copies of the images (256×256 pixels) contained all the distinguishing features relevant to characterising IEC. Importantly, each patch was rotated so that the epidermis was horizontal, resulting in a fixed and consistent perspective of the stratified layers. The patches were curated so that out-of-focus patches, as well as patches on the edge of specimens were removed. The resulting dataset consisted of 8,490 images at 256×256-pixel resolution, representing variations of tissue morphology within the keratin, epidermal and papillary dermal layers, as well as physical variations such as orientation ink, staining and lighting. Augmented with a left-right flip, the final dataset contained 16,980 images.

## Natural Language Annotations

The fixed context window enabled a systematic approach to annotating the images to be followed. In consultation with a pathologist, we created a controlled vocabulary to characterise each image according to the predictable locations of tissue types. The vocabulary included descriptors at a high but clinically relevant detail, providing a succinct yet accurate summary of the three tissue types. For the keratin, epidermal and dermal layer, the descriptions were broken down as follows:

Keratin:
*The upper layer appears/shows (thin/thick/fragmented/detached)*
- *basket weave keratosis*
- *basket weave keratosis with parakeratosis*
- *parakeratosis*
- *keratosis*
- *eroded*

Epidermis:
*The epidermis appears/shows*
- *normal*
- *mild/moderate/severe/full-thickness dysplasia*

Dermis:
> *The dermis appears/shows/has been*
> - *abnormal*
> - *normal*
> - *solar damaged*
> - *inflammation*
> - *displaced*

At the level of detail in the description, it was decided that left-right flips of images could be accurately described using the same caption. Thus, all images in the dataset were associated with a natural language caption between the length of 14-24 words, composed of three sentences e.g.

> *The upper layer shows thick and fragmented parakeratosis. The epidermis shows severe dysplasia. The dermis shows inflammation.*

The result is a first of its kind dataset for histological images of skin cancer with 8,490 unique natural language descriptions of tissue morphology for 16,980 images.

## Image Reconstruction

Following the original VQ-GAN architecture, we implemented our own version using Tensorflow 2.5 [32]. A high-level overview of the architecture is shown in **Fig. 1**. The encoder and decoder design included the original Residual Blocks with Group Normalization, Dropout and Swish activation, as well as Spatial Attention. The encoder, quantizer and decoder components constitute the original VQ-VAE formulation [33]. The encoder takes an image *x* e.g., 256×256×3, and maps it to $z_e$, a 16×16×$d$ feature representation. This is fed to the quantizer which calculates the distance between $z_e$ and a collection of embedding vectors $e$, of size $d$, and $k$ is the total number of vectors. The index of the embedding with the shortest distance to the individual features of $z_e$, constitutes the discrete representation of the image, $q$ e.g., 16 ×16, found by:

$$argmin\left[\Delta(z_e, e)\right] = argmin_j \|z_e - e_j\|_2$$

The corresponding embeddings form $z_q$, from which the decoder outputs a reconstruction of the original image, $x'$. The embedding loss attempts to move the embedding vectors towards $z_e$ via the L2 loss. A second term is added to the embedding loss, referred to as the commitment loss, which ensures the encoding does not grow arbitrarily and the encoder commits to one of the embedding vectors. A commitment coefficient β is included, which in all our experiments had β=1. The embedding loss is therefore defined as:

$$\text{Embedding Loss} = ||sg[z_e] - e||_2^2 + \beta ||z_e - sg[e]||_2^2$$

where $sg$ is the stop gradient operation. The reconstruction loss is composed of an L1 loss term between $x$ and $x'$, a perceptual loss term and an adversarial loss term. Perceptual loss is defined as the L2 distance between features of the inputs $x$ and $x'$, in VGG16 perceptual space [34]. The adversarial loss is implemented as a PatchGAN [35], which classifies small patches of the input images $x$ and $x'$, as either real or fake. The combined effect is that the L1 loss optimises global structure, and the perceptual loss and adversarial loss optimize for perceptually similar and realistic features. Unlike in the original VQ-GAN which dynamically weighted the perceptual and adversarial terms, our implementation achieved successful results with all terms weighted equally, thus:

$$\text{Reconstruction Loss} = L1(x, x') + \text{Perceptual Loss}(x, x') + \text{Adversarial Loss}(x, x')$$

Various parameterisations of the model were explored in terms of embedding dimensions *d*, the number of embeddings *k*, the down sample factors *f*, and the number of filters. Our best results, in terms of reconstruction quality, were achieved with $d = 256, k = 2048, f = 4$, and with filters for each subsequent block in both the encoder and decoder as: 256×256 →64, 128×128 →128, 64×64 →256, 32×32→ 512, 16×16 → 512. An Adam optimizer was used with a learning rate of $2 \cdot 10^{-5}$, with $\beta_1 = 0.5$, and $\beta_2 = 0.9$. A non-saturating adversarial loss was used, with the parameters of the PatchGAN following the original implementation. Using a 64:18:18% training, validation and test split, the model was trained up until the validation embedding loss converged which was around 130 epochs. We used a batch size of 24 and utilized two NVIDIA SXM2 Tesla 32GB V100 GPUs.

The code for all the models can be found at
https://github.com/smthomas-sci/SkinCancerLanguageCharacterisation

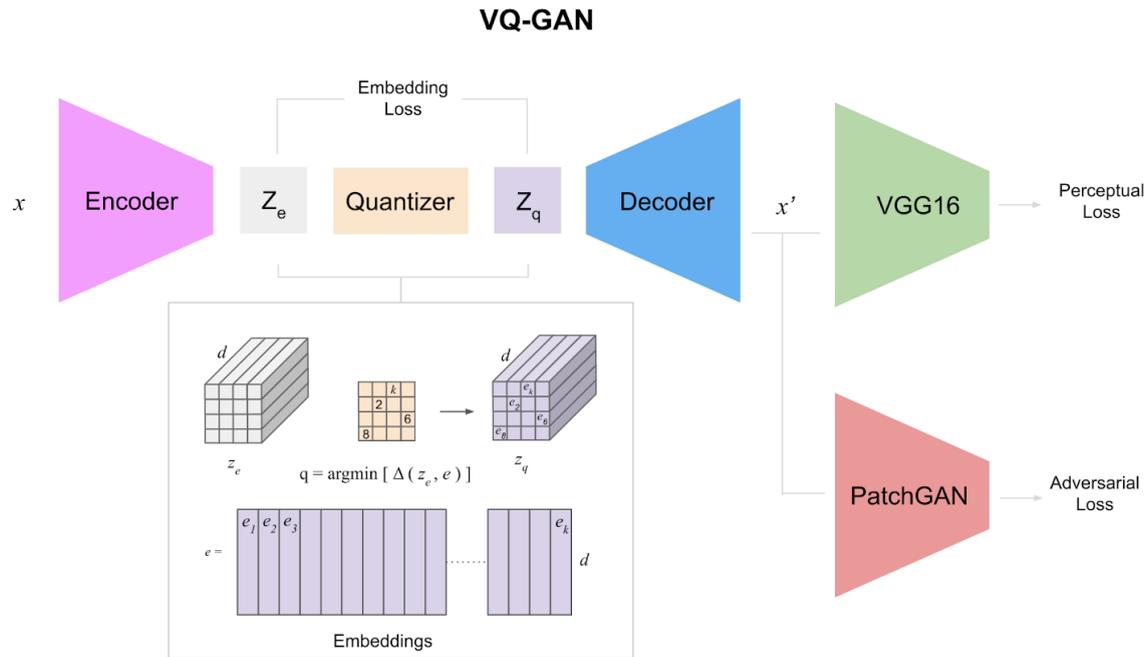

**Fig. 1: The VQ-GAN is a fully convolutional network which contains three main sub-networks: the encoder, quantizer and decoder.** The encoder takes an input image *x* (256×256×3) and transforms it down to $z_e$ (16×16×256). $z_e$ is fed to the quantizer which selects the index of the closest embedding *e* to produce a discretized representation, *q*. The indices of *q* are then used to select the embeddings to form $z_q$ from which the decoder reconstructs an image $x'$. The reconstruction quality is driven by the perceptual loss and adversarial loss, which encourages realistic features to be modelled. The embedding loss moves the embedding vectors towards $z_e$.

## Autoencoding Features

The VQ-GAN is a fully convolutional network which means the lowest feature representation $z_e$ retains spatial structure. However, other generative methods such as GANs encode the input domain into a flattened *n*-dimensional space, which projects the spatial and feature information into a single encoding. This enables interpolation through the space to visualise its structure, as well as the use of concept vectors as an interpretability method [18, 24, 31]. We therefore wanted to extend the generative capacity of the VQ-GAN by combining it with more traditional autoencoding methods. Our idea was to autoencode the rich 16×16×256 $z_e$ features using only L2 loss. This is possible because optimising in feature space can lead to better results than optimizing pixel space directly i.e., perceptual loss and inversion networks [36, 37]. We therefore create a convolutional encoder and decoder which maps from 16×16×256 to 4×4×1024 and back, with a flattened central latent space *w* of 512-dimensions (**Fig. 2**). We did not include latent space constraints such as independence of dimensions such as in a VAE since we were not interested in sampling. Although foreseeably this approach might impact the smoothness of the latent space, it did not have a significant effect in practice. Consequently, this architecture allowed us to manipulate the *w* representation of an image, reconstruct the $z_e$ representation, and subsequently pass that to the quanitizer and decoder to obtain the corresponding image. This ties together both the benefits of traditional latent space generative modelling techniques as well as discrete representation learning.

We trained the network on the same training and validation set using the $z_e$ features from VQ-GAN for 300 epochs until the validation loss plateaued. We used an Adam optimiser with a learning rate of 0.001 and $β_1$=0.9, and $β_2$=0.999.

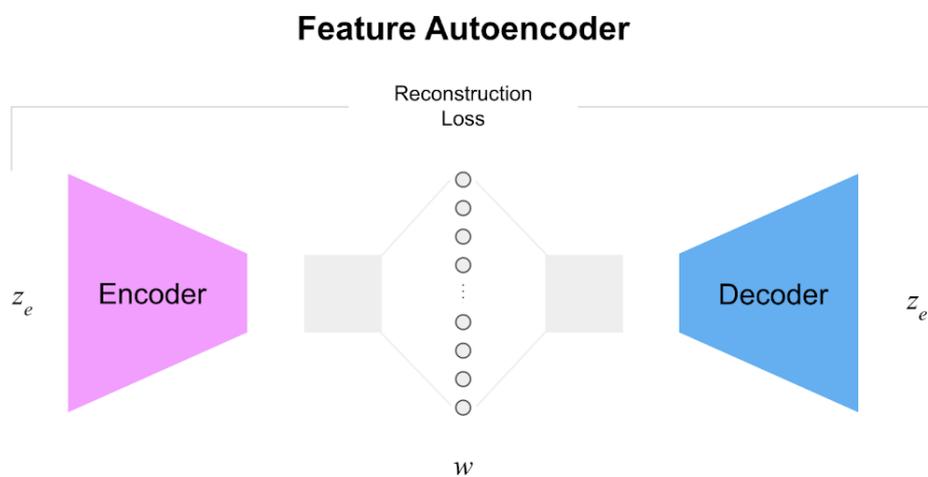

**Fig. 2: The Feature Autoencoder is standard convolutional autoencoder with a flatten bottleneck.** The network takes $z_e$ (16×16×256) and after two convolutional blocks, flattens a 4×4×1024 spatial feature space into *w*, a 512-dimensional latent space. The decoder then reconstructs $z_e$ using mean squared error. There is no regularization of the latent space for smoothness, such as in GANs and VAEs.

## Image Captioning Technique

The problem of image captioning attempts to provide a relevant natural language output which summarises the content of an image. In mapping from the image domain to the text domain, we conceived of the tasks as a sequence-to-sequence problem. To this end, we implemented a single-head encoder-decoder transformer model (**Fig. 3**) which takes the flattened embedding indices *q*, encodes it, and then decodes it into a text sequence. We again implemented our own model in Tensorflow 2.5 following the original model formulation [15]. After a parameter search our final model consisted of 4 single-head attention blocks with 16-feed-forward nodes in each block. The image domain vocabulary size was the same as the number of discrete image features, $k = 2048$. The text vocabulary size was 250, after running the BERT word-piece tokenization algorithm [38]. We trained the transformer model using an Adam optimiser with a learning rate of 0.001, $\beta_1$=0.9, $\beta_2$=0.999, and a drop-out rate of 0.1, minimising the categorical cross-entropy loss for predicting the next token in the sequence. The model was updated over 50 epochs and we selected the weights from the best validation loss. We found performance to be similar across many parameter choices and so opted for the smallest high-performing model.

To generate captions with the trained model, the image sequence was encoded and fed to the decoder which was conditioned with a "[START]" token. The decoder generated the next token using argmax sampling, which was concatenated to be fed back into the decoder. The sequence was iteratively generated until a "[STOP]" token was generated.

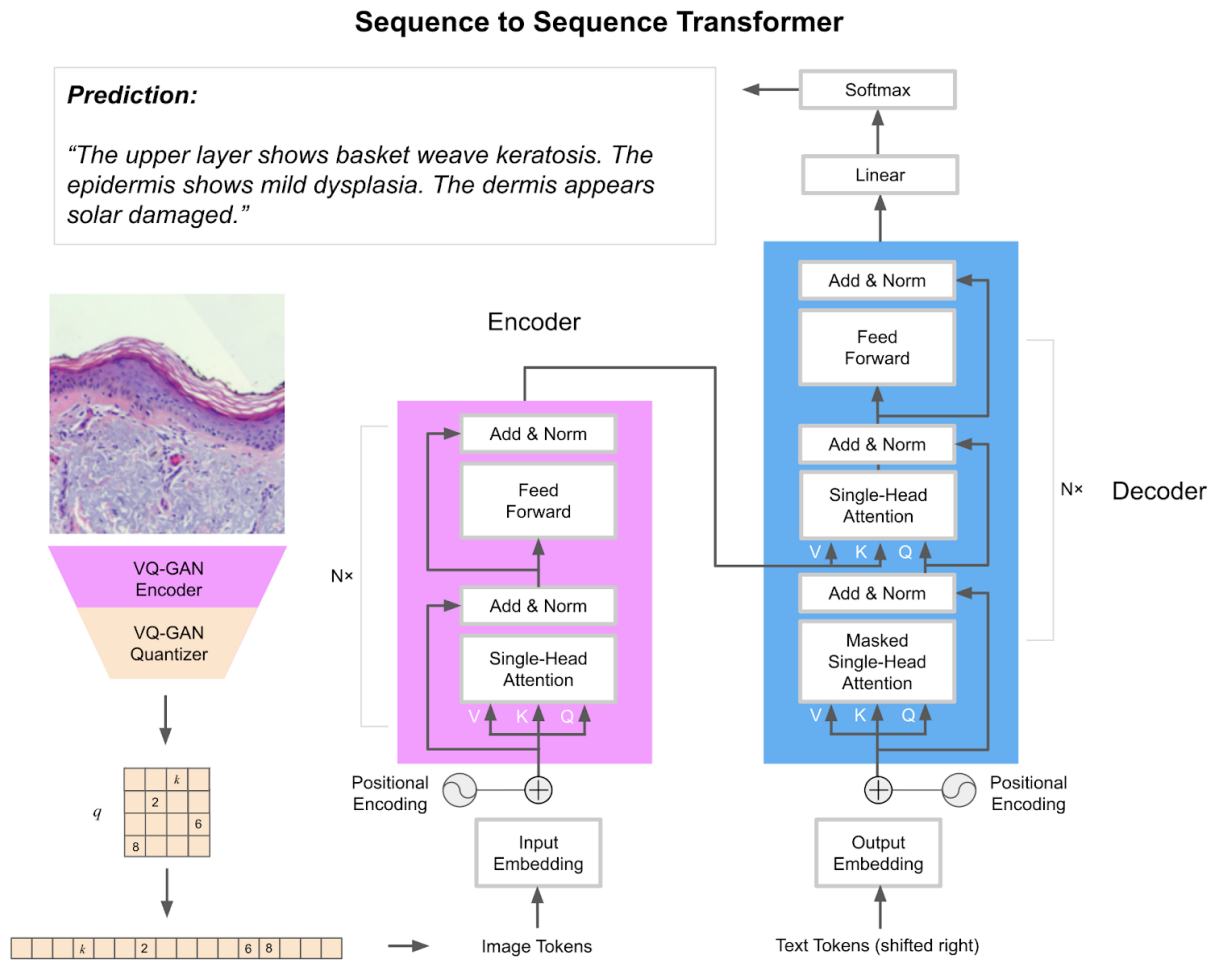

**Fig. 3: A sequence to sequence transformer model is used to map between the discrete image domain and the discrete natural language domain.** Both the encoder and decoder components have the same parameterization, specifically single-head attention with 16 feed-forward nodes and 4 layers (N) each. The image domain and text domain vocabulary sizes are 2048 and 250, respectively. The whole pipeline consists of passing an image through the VQ-GAN encoder and quantizer, flattening the discrete representation and then passing it through the transformer encoder.

# Results and Discussion

## Image Reconstructions from Discrete and Continuous Space

One theoretical benefit of the VQ-GAN model is that it retains spatial information and can thus focus on learning the local tissue features directly. The image reconstructions in **Fig. 4** indicate that the network has learned a diverse set of embedding features, which can be reasonably reconstructed at the 256×256-pixel scale. The quality of the reconstructions compared with the original training data (11,588 images) was measured using the Frechet Inception Distance (FID) [39], resulting in a score of 17.6. As this is the first time using this dataset, the number is only a benchmark (see later), and the quality of the reconstructions can only be determined qualitatively. From that perspective, the global structure of the images is very good, covering much of the major variation in the three major tissue types. However, there are important details still lacking in the epidermal layer, particularly the sizes and color variations (dysplasia) of the keratinocytes. It appears that the features generated are smoothed versions of the original features or are replaced with more common cellular features. In contrast, the model appears to accurately capture most features in the keratin and dermal layers.

As learning dysplastic features is a key aim of the work, effort was made to understand why such features were lacking. The first approach was to explore different embedding sizes, varying both d, the depth, and k, the number of embeddings. We obtained similar results for $d \in \{256, 512\}$, and found results improved as *k* increased, with 2048 being the best. This was in line with the original work that found increased *k* produced better results [12] . Interestingly, when looking at the unique embedding indices utilized across the whole training dataset, we found that only 291 indices were used, despite another 1757 being available. We hypothesize that the individual embeddings form atomic features which can be combined to produce most other features.

We show in the supporting results **Fig. S1** comparisons between images generated from random *q* representations from the utilized indices and non-utilized indices. Importantly, the non-utilized indices also produce realistic images, albeit of a less diverse nature, indicating that most embeddings were optimized over the course of training. Indeed, the idea that atomic embeddings are combined to produce realistic features is further demonstrated in **Fig. S2**, which shows the Top 25 most common indices, and the corresponding generated image. They highlight key variations in background and dermal textures and colours, but by themselves lack much of the realism compared to the combination of indices. These results suggest that the size of the embedding space was not a significant limitation in attempting to generate the full space of features.

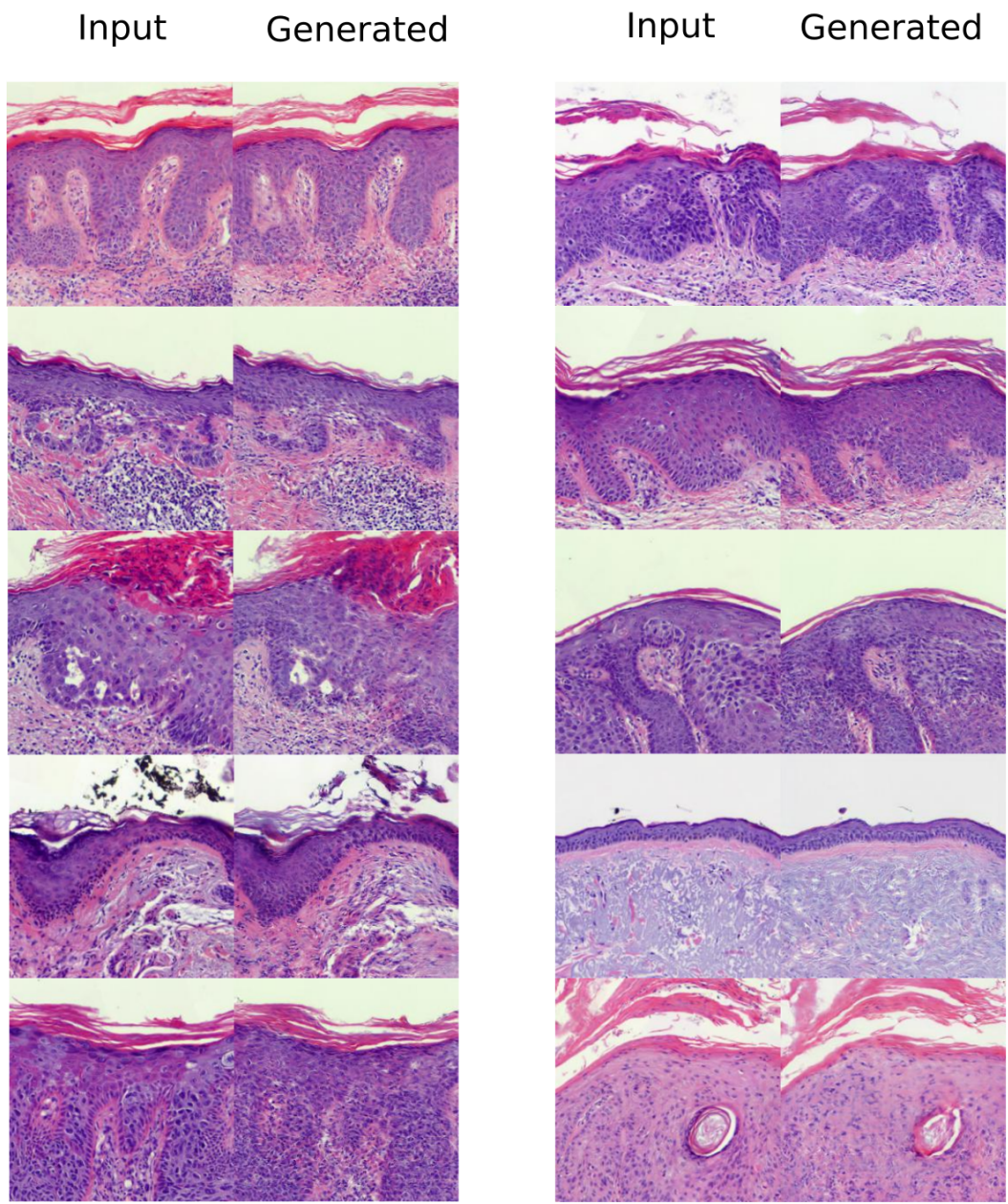

**Fig. 4: Image reconstructions from the VQ-GAN model.** Unseen test-set reconstructions for images from the VQ-GAN model showing that global context and diverse tissue features can be successfully regenerated at the 256×256-pixel level.

    The second idea was whether the PatchGAN had a big enough receptive field to capture the size of the cellular features in question. We calculated the receptive field to be 72×72 pixels, which was sufficient to cover the size of keratinocytes. To verify, we experimented with the receptive field size using atrous convolutions in the PatchGAN, as well as changing the down-sampling factor to 3 (resulting in a *q* of size 72×72). In each case, keratin and dermal layers remained similar, with no improvement for the epidermal features. Thus, in general, it is clear that PatchGAN is able to direct the network to generate realistic features and may not be the cause of this issue.

We thirdly thought that the networks preference for some features over others may be due to their relative frequency. We hypothesize that the perceptual loss component directs the network to learn the most frequent features in a region, rather than all specific features. It is observed in other perceptual loss reconstruction problems that networks learn to generate spurious patterns to satisfy the objective on complex regions of the image, such as human hair [40]. We therefore suspect that since the placement of pixels is indirectly penalized, the network attempts to generate features which are, in general, characteristic of a local region of the image. If this is the case, it makes sense for the adversarial and perceptual loss terms to be uniquely weighted, either dynamically as in the original VQ-GAN formulation, or as a fixed hyperparameter. Despite this, the reconstructions are spatially accurate and texturally diverse, demonstrating for the first time high-quality and high-resolution reconstructions of skin cancer images.

The reconstructions from the feature autoencoder are shown in **Fig. 5**. Compared again to the training set, the FID score of the reconstructions was 36.1, indicating a greater distance to the original images than the base VQ-GAN. Visual inspection of the images confirmed this. Interestingly, the latent space encodes an enormous variety of spatial information, while still capturing reasonably well realistic features. The more simplified and repetitive epidermal features are more pronounced in these images, tending towards textual similarity rather than specific features. This could again be a consequence of learning average features via the L2 loss in feature space, in this case having a more pronounced effect than the official perceptual loss. However, the quantization component provides the added advantage that despite not being able to faithfully reconstruct the original image, the nearest realistic feature is utilized in the reconstruction. This straightforward approach would not be possible using a standard autoencoder, which tends towards blurry images. In general, the *w* vector captures both spatial and textural information in a meaningful way and to a quality that does not impede latent space exploration (see *Latent Space Exploration* section).

Interestingly, the number of indices utilized for the training set was 380, sharing 282 codes with the original $z_e$ quantization. Although there is a reduction in quality, the small number of utilized indices indicates further that most indices are optimized towards useful features. Evidence of this is seen in the supporting results **Fig. S3**, which shows reconstructed images from the nearest 4 indices, all generating realistic, but decreasing quality reconstructions. It may be that once on a training path, the network converges on similar competing features, rather than diversifying its weights to capture additional ones. As mentioned above, this might be overcome with a stronger signal from the adversarial loss term.

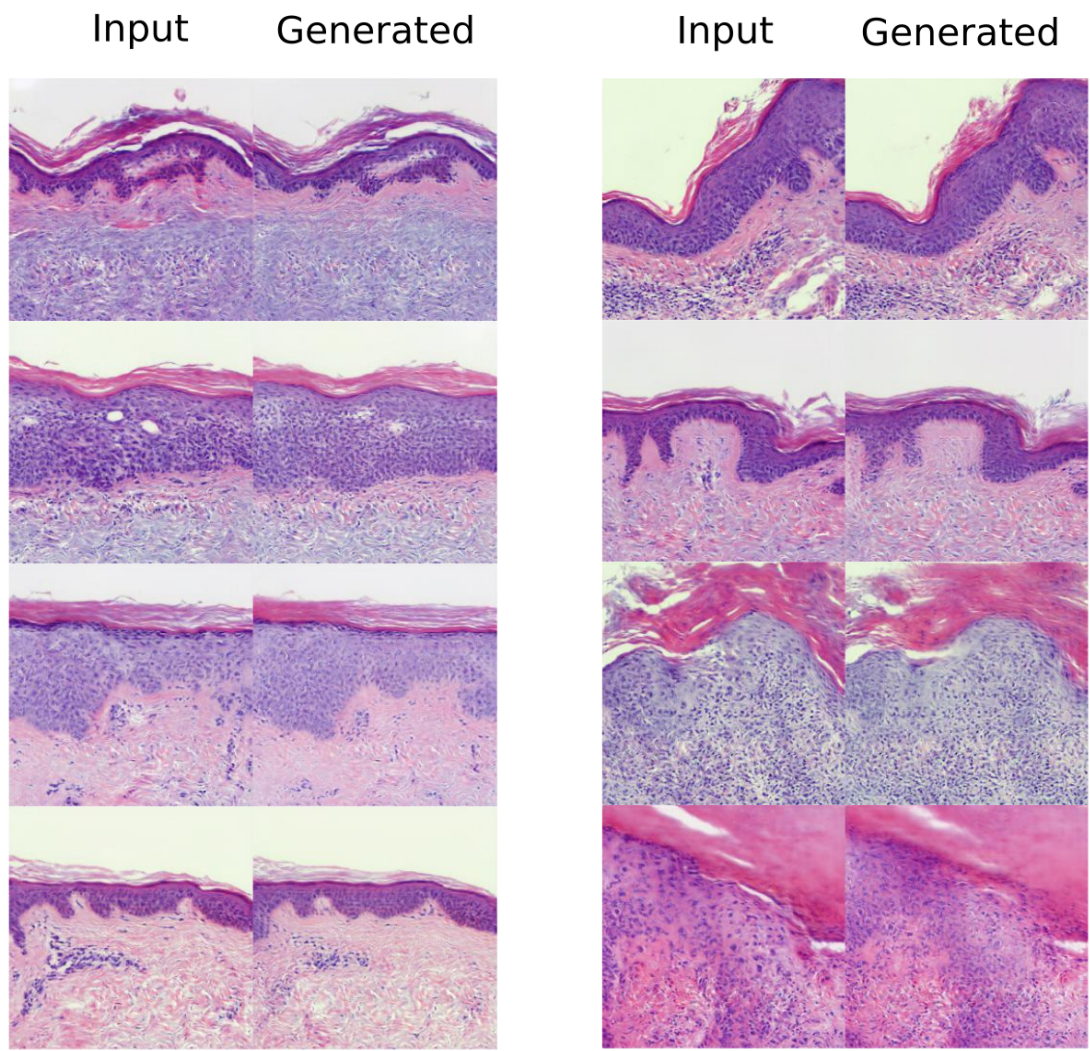

**Fig. 5: Image reconstructions from the Feature Autoencoder model fed to the VQ-GAN.** Unseen test-set reconstructions (256×256) shows that the Autoencoder model can successfully project 16×16×256 features into a flattened space (*w*) while retaining a substantial amount of rich spatial and feature information.

## Image Captioning

The sequence-to-sequence transformer model proved to be an effective way to model the controlled vocabulary of the pathologist. Examples of captions for a variety of test set images are shown in **Fig. 6**. During training, the cross-entropy loss penalised the predictions for each position in the sequence, and so the inclusion or omission of modifiers such as *thin/thick/fragmented* etc. were classified as incorrect, despite the *object* of the modifier being correct, e.g., *basket weave keratosis*. However, when we compare the ground truth labels and predictions alongside the images, we get a general sense that the model performs reasonably well in this controlled environment. The major pattern is as just described, namely that the errors are largely from the inclusion or omission of modifiers, and instances where the decision boundary is arbitrary e.g., both solar damage and inflammation are present in the dermis. In some cases, there is no clear semantic difference in using keratosis over basket weave keratosis / parakeratosis and this is just the result of the original annotation.

Performance metrics for natural language tasks tend to give an incomplete picture of the quality of the model [41]. In our case, we assessed the model based on one-to-one comparisons between predicted words and the ground truth, penalising partial matches as done in training. Despite this strict criterion, we achieved high and consistent accuracies across the training, validation and test sets (same sets as image reconstruction), with scores of 96%, 93% and 92% respectively. These numbers seem to reflect the quality observed in **Fig. 6**.

In exploring various hyper-parameters, ranging from 60,000 weights to 160,000 weights, the performance remained the same suggesting that the small output vocabulary provided a strong signal. This may also be owed to the additional fact that the training set only used 291 indices. Consequently, this indicates that the input vocabulary of 2048 provided redundant weights, since the unused inputs are effectively zeros. However, since the weights are only in the input/text embedding layer, the extra weights could not cause overfitting.

The advantage of an ideal combination of discrete and continuous methods is that we can generate an image caption, manipulate the image via concept vectors (next section), and then generate a new image caption. In this way, we have both visual and language correlations to inspect the quality of *both* models. With an increasing vocabulary and greater coverage of the image domain, we can envisage a system which could become a distillation of the current best knowledge for this problem domain. However, obtaining the flexibility and richness of natural language provides a major challenge, specifically with deciding on the extent to which image features should be described, as well as then scaling the dataset to make training such a model possible. Therefore, at this stage, the problem formulation is more suitable as an interpretability method rather than being directly incorporated into a clinical report. Provisioning for open-ended evidence gathering to serve the ultimate task of diagnosis remains an open problem, with this work serving as the first step in that direction.

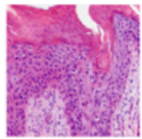

**GT**: The upper layer shows parakeratosis. The epidmeris shows full-thickness dysplasia. The dermis has been displaced.

**P**: The upper layer shows thick parakeratosis. The epidmeris shows full-thickness dysplasia. The dermis shows inflammation.

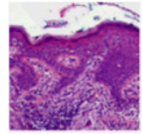

**GT**: The upper layer shows keratosis. The epidmeris shows moderate dysplasia. The dermis shows inflammation.

**P**: The upper layer shows thin keratosis. The epidmeris shows severe dysplasia. The dermis shows inflammation.

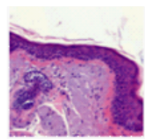

**GT**: The upper layer shows fragmented keratosis. The epidmeris shows moderate dysplasia. The dermis appears solar damaged.

**P**: The upper layer shows basket weave keratosis. The epidmeris shows mild dysplasia. The dermis appears solar damaged.

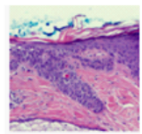

**GT**: The upper layer shows basket weave keratosis. The epidmeris shows severe dysplasia. The dermis appears normal.

**P**: The upper layer shows basket weave keratosis. The epidmeris shows severe dysplasia. The dermis appears normal.

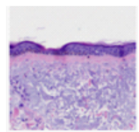

**GT**: The upper layer shows thin basket weave keratosis. The epidmeris shows mild dysplasia. The dermis appears solar damaged.

**P**: The upper layer shows thin basket weave keratosis. The epidmeris displays shows mild dysplasia. The dermis appears solar damaged.

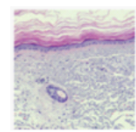

**GT**: The upper layer shows basket weave keratosis. The epidmeris appears normal. The dermis appears solar damaged.

**P**: The upper layer shows basket weave keratosis. The epidmeris shows mild dysplasia. The dermis appears solar damaged.

**Fig. 6: Sequence to sequence modelling results in accurate and naturally interpretable characterisations of the test image content using natural language.** The variation in results indicates the model has correctly learned the independence of the keratin, epidermal and dermal layers. In some cases, the prediction differs from the ground truth, but is actually an accurate description e.g., predicted "inflammation" instead of "solar damaged" despite both features being present in the image. The above images are hand-selected samples from the test set to represent the diversity of tissue morphology and corresponding annotations.

## Latent Space Exploration

The accompanying annotations for each image allowed five meaningful concept vectors to be defined, specifically: keratin thickness, parakeratosis, dysplasia, solar damage, and inflammation. Following the method in [18] they were defined using a binary linear classifier using subsets of the data showing prototypical and contrastive features of each concept e.g., thin versus thick keratin and normal versus solar damaged dermis. The classification accuracies are seen in **Table 1**, showing that a high degree of accuracy was achieved for most concepts across the training, validation, and test data. The lower sensitivity scores for normal dermis (for both the solar damage and inflammation concepts) are likely explained by the small number of positive normal dermis examples, as well as the greater imbalance between negative examples. However, this did not seem to affect the quality of concept interpolations (see later). The normalized coefficients and biases of the respective linear classifiers were then used to define the concept vectors and explore the latent space.

**Table 1: Concept vector binary classification scores.** These scores are produced from a binary classifier trained to predict positive (+) and negative (-) concept classes with *n* number of images. The classifier takes the *w* representation of the image, where *w* is of size 512. The normalized coefficients and bias terms from each linear classifier are then used to define the respective concept vectors, each denoting a direction in *w*-space.

| Concept | Binary Classes | Data Set | + n | - n | Accuracy | Sensitivity | Specificity |
|---|---|---|---|---|---|---|---|
| Keratin thickness | Thin vs Thick keratosis | Train | 1684 | 1996 | 0.9989 | 0.9988 | 0.9990 |
| | | Validation | 396 | 480 | 0.9600 | 0.9621 | 0.9583 |
| | | Test | 370 | 446 | 0.9608 | 0.9838 | 0.9417 |
| Parakeratosis | Basket weave vs Parakeratosis | Train | 6230 | 4244 | 0.8513 | 0.8777 | 0.8124 |
| | | Validation | 1452 | 990 | 0.8448 | 0.8533 | 0.8323 |
| | | Test | 1506 | 886 | 0.8106 | 0.8386 | 0.7630 |
| Dysplasia | Mild dysplasia vs Full thickness | Train | 2704 | 4536 | 1.0 | 1.0 | 1.0 |
| | | Validation | 868 | 1146 | 0.9230 | 0.8917 | 0.9468 |
| | | Test | 592 | 1030 | 0.9322 | 0.9071 | 0.9466 |
| Solar Damage | Normal dermis vs Solar damage | Train | 1932 | 3544 | 0.8486 | 0.7629 | 0.8953 |
| | | Validation | 374 | 1102 | 0.8388 | 0.6310 | 0.9093 |
| | | Test | 340 | 904 | 0.8223 | 0.7059 | 0.8662 |
| Inflammation | Normal dermis vs Inflammation | Train | 1932 | 3176 | 0.8955 | 0.8540 | 0.9207 |
| | | Validation | 374 | 608 | 0.8065 | 0.6364 | 0.9112 |
| | | Test | 340 | 718 | 0.8270 | 0.6471 | 0.9123 |

The 2-dimensional UMAP projections of the *w* representations of 11,588 training images are shown in **Fig. 7**. The annotation categories used to define the concept vectors are overlaid for each respective tissue layer in **Fig. 7a**. On first inspection, it indicates that opposing/contrastive concepts in general are allocated to different regions of the latent space. Looking across all tissue layers indicates further the independence of some concepts, e.g., inflammation with both thin and thick keratin, as well as correlations of others e.g., thick keratin with parakeratosis. Interestingly, the labels provide an indication that the latent space captures the progression pattern of mild to severe dysplasia, with full-thickness dysplasia placed in polar contrast to mild. Although the details of the progression are not clearly captured in the image reconstructions, the information may be persevered to a reasonable degree in the $z_e$ feature space and the encoder of the autoencoder. Therefore, these labels hint that the latent space structure conforms in some way to our biological expectations.

Example images from different regions of the latent space (**Fig. 7b**) further demonstrate how features are arranged according to their high-level meaning. This is in terms of both the biological e.g., tissue type variation, as well as histochemical, such as dermal staining colours and intensity. Of note is the outlying group, presenting abnormal keratin, epidermal and dermal layers, with no smooth semantic connection to the other images. In contrast, the smoothness between different regions of the latent space is shown in **Fig. 8**. Linear interpolations between two points in the latent space show a smooth semantic transition, following a natural journey. Specifically, the keratin layers transition smoothly e.g., from thin to thick, and the epidermis and dermal regions transform without out-of-context intermediary images, which is noted in previous work [18, 31]. Of note are the interpolation journeys through the UMAP projection space, emphasizing that steps and directions in 512-dimensional space are not fully represented in 2 dimensions due to information loss. However, the traversals through the high-dimensional latent space and their corresponding visualisations are a compelling means of testing hypotheses about the quality of the representations and their overall structure (see next).

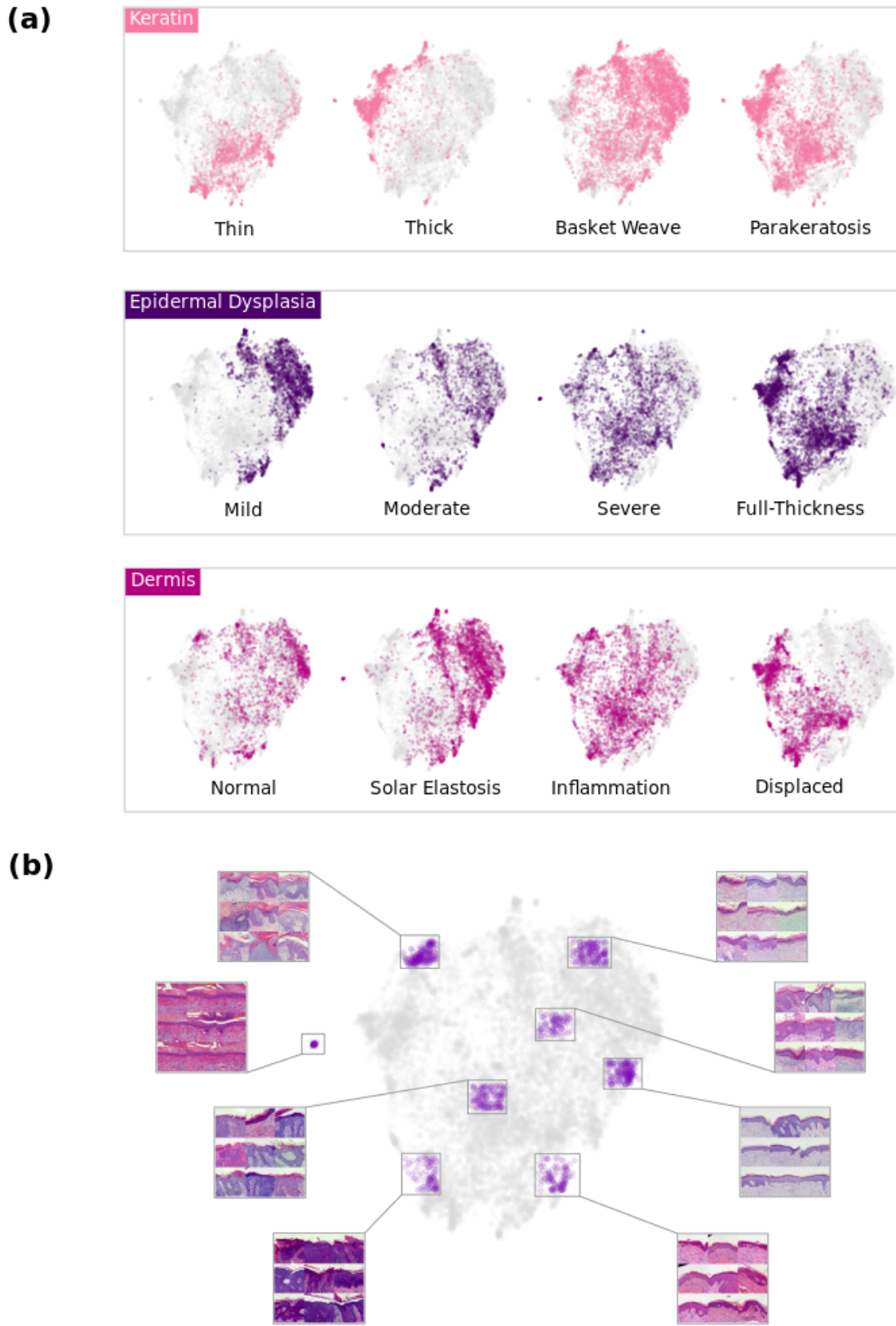

**Fig. 7: Projections of 11,000 training images representations ($w$), from 512 dimensions to 2 using the UMAP algorithm.** (a) Various labels relevant for each tissue layer extracted from the annotations indicate that similar text descriptions correspond to similar regions of the latent space. (b) Examples of images from different regions of the latent space show separation between mildly and severely dysplastic images, thickness of the keratin layer and depth of invasion, as well as variations in staining.

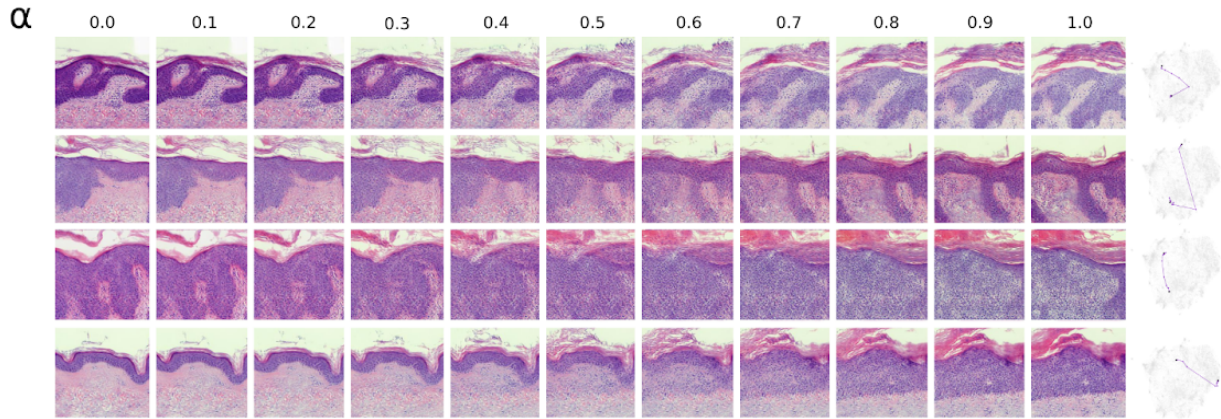

**Fig. 8: Linear interpolations between image pairs in the latent space.** The results show realistic journeys between endpoints where features transition in a smooth manner with intermediate images presenting realistically blended features and spatial context of both endpoints.

Manipulating images using the learned concept vectors generated exciting results (**Fig. 9**), demonstrating for the first time the ability to systematically manipulate skin cancer images in a desired and biologically meaningful way. Each row shows the transversal through the latent space (*w*) in a fixed direction defined by the concept vector. There are two major results observed. The first is that the additive or subtractive operation corresponds to the increase or decrease of the relevant features for each concept. Moving from left to right we can see that (1) keratin transitions smoothly from thin to thick, (2) basket weave keratosis transforms into thicker parakeratosis, and (3) the dermal regions begin to include more solar damage or inflammation. (We will discuss the transition from mild to full-thickness dysplasia shortly). The second major result is that for those mentioned concepts, there is a clear independence of the relevant features and the remaining image content. That is, applying a concept vector only affects those features pertaining to the concept, with other features remaining the same. This is seen very clearly for solar damage and inflammation, where we see the dermis change but the epidermis and keratin layers remain the same. This is largely true for thin to thick keratosis, as well as basket weave to parakeratosis, perhaps with the inclusion of correlated features such as inflammation. Indeed, the fact that correlated features are included, despite not being directly used as selection criteria is precisely the value of this technique. The ability to do this is a major milestone towards using unsupervised learning to capture the subtleties of biomedical imaging, and then learning about them ourselves in a post-hoc manner. We believe that the major contributing factor to the smoothness of the transitions and natural journeys seen in **Fig. 8** and **Fig. 9**, is that care was taken to orientate the images, and create a fixed perspective of the problem domain.

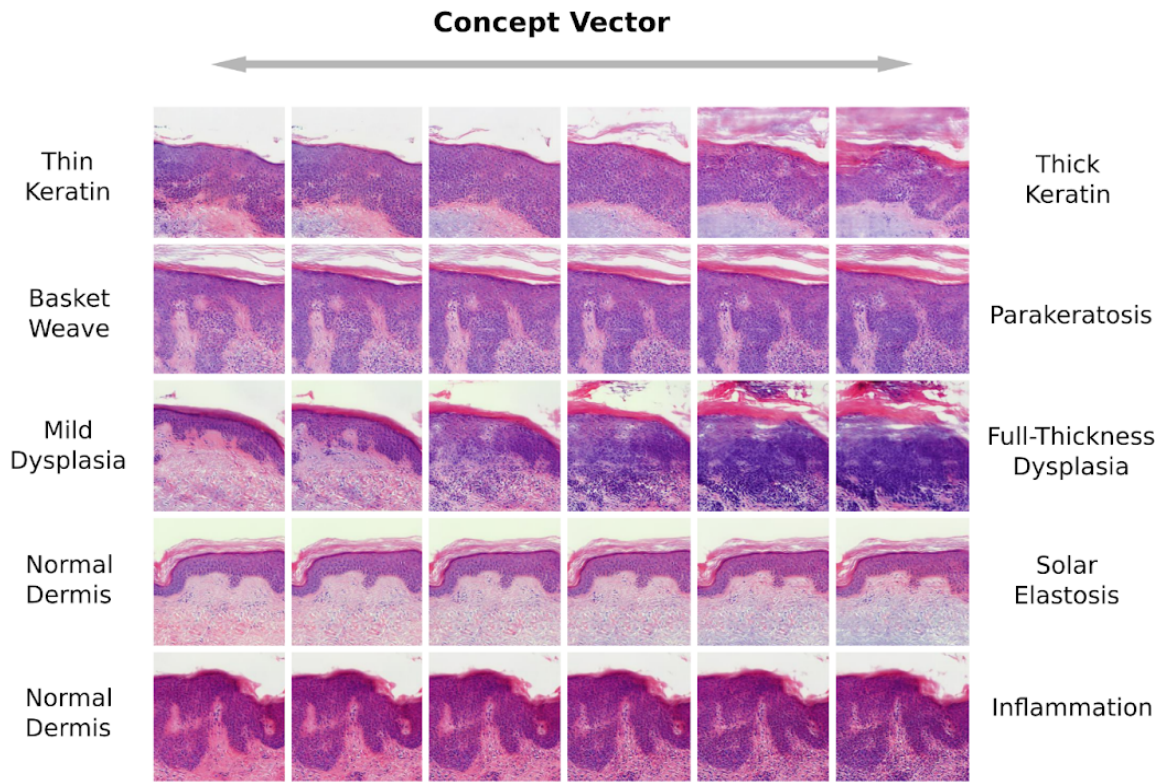

**Fig. 9: Interpolations for concept vectors related to the keratin, epidermal and dermal layers of the skin.** The results indicate that using concept vectors as an interpretability method is now possible across a variety of high-level and subtle concepts. Concepts appear to transition smoothly, with correlated features changing and independent features remaining constant.

The transition from mild to full-thickness dysplasia presents a more challenging problem. As discussed previously, visualising the key features of IEC proved difficult. Therefore, the concept vector does not provide a clear insight to how the epidermal cells transform. However, moving from left to right, we do see correlated features change, with the thickening of the keratin layer and inclusion of parakeratosis, the thickening of the epidermis and the introduction of inflammation. At the very least, the epidermis does show a deviation from normal towards a more disorganized and aggressive appearance. Therefore, these transitions are indicative that the concept vectors include relevant features, and what is currently limited is the ability to adequately visualise the information. Knowing what can now be achieved, the focus on future work needs to be on improving the image quality of the epidermal cells. The obvious advantage to this technique is that our ability to understand the workings of both the model and problem domain increases as the quality of visualisations improves.

# Conclusion

Our aim was to use machine learning to model a key aspect of the pathologist workflow, specifically, using natural language to describe complex features of cellular morphology. We created a dataset of high-resolution images (256×256 pixels) of intraepidermal skin cancer, which were paired with natural language annotations using a controlled vocabulary of pathology terms. The method involved combining discrete and continuous generative modelling techniques in a unique way, with the goal of illuminating the continuous progression of disease states from healthy to cancerous. Using a VQ-GAN, high-quality reconstructions of a diverse set of images were achieved, while additionally learning feature-rich low-dimensional representations. Pairing the discrete image representations with the natural language annotations, we trained a sequence-to-sequence transformer model to generate accurate and realistic image captions across the full diversity of images. Additionally, we used an autoencoder on the VQ-GAN features to learn a continuous representation, which resulted in a smooth and semantically meaningful latent space useful for model interpretability. Importantly, we show for the first time the ability to meaningfully manipulate images of skin cancer using concept vectors defined from image captions. Ultimately this work serves as a valuable demonstration of the power of text and image generative models to aid our understanding of diseases such as non-melanoma skin cancer.

**Conflicts of Interest**

The authors declare that they have no known competing financial interest or personal relationships that could have appeared to influence the work reported in this paper.

**CRediT authorship contribution statement**

**Simon M. Thomas**: Conceptualization, Methodology, Software, Formal analysis, Investigation, Data curation, Writing - original draft, Visualization. **James G. Lefevre**: Conceptualization, Supervision, Writing - review & editing. **Glenn Baxter:** Conceptualization, Investigation, Resources, Supervision, Writing - review & editing. **Nicholas A. Hamilton**: Conceptualization, Supervision, Writing - review & editing, Resources, Project administration.

**Acknowledgements**
We wish to acknowledge The University of Queensland's Research Computing Centre (RCC) for its support in this research. JL is supported by Australian Research Council Discovery Grant DP180101910. MyLab Pathology provided access to their archived histological collection. Financial support was provided by The Australian Government Research Training Program (RTP) and The Laurel Joy George Perpetual Scholarship.

Supporting Results

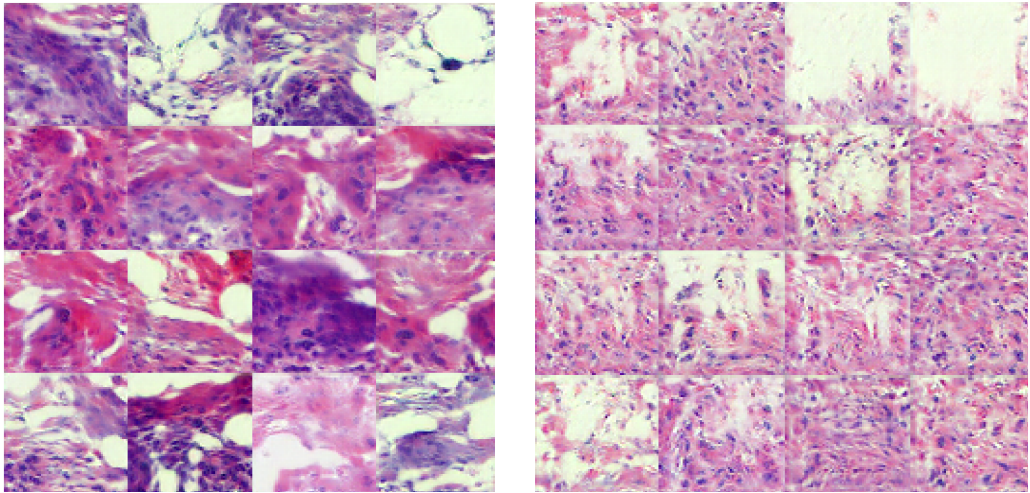

**Fig. S1: Texture generation using random embedding indices.** Each image corresponds to a *q* of size 4×4, which leads to an image of size 64×64 pixels. Left) Examples of images generated from random samples from the 291-code vocabulary. Right) Examples of images generated from random samples from the out-group codes (1757). The utilized codes (left) express more variation in the textures and appear realistic. The outgroup codes, although of realistic appearance, are more uniform in texture and the variety of features.

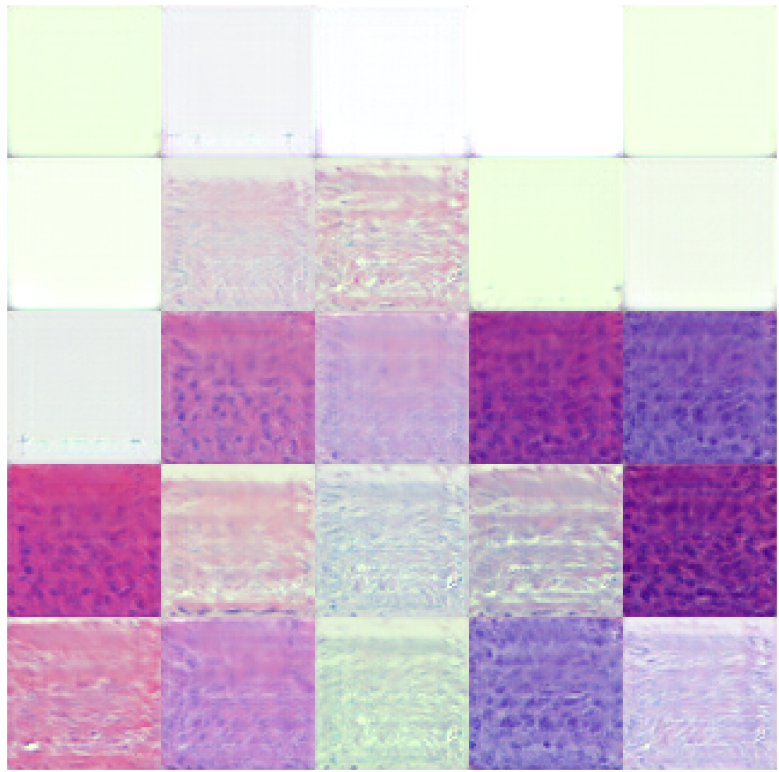

**Fig. S2: Texture generation of the top 25 embedding indices.** Each image corresponds to a *q* of size 4×4, which leads to an image of size 64×64 pixels. The images are ranked from 1-25, firstly across columns and then rows. For example, column 2, row 2 corresponds to image number 7. Background codes are the most common, with dermal textures and stain variations also clearly seen. There is a lesser sense of realism to these images, indicating that realistic features are achieved by combining codes in various arrangements, as seen in the previous figure.

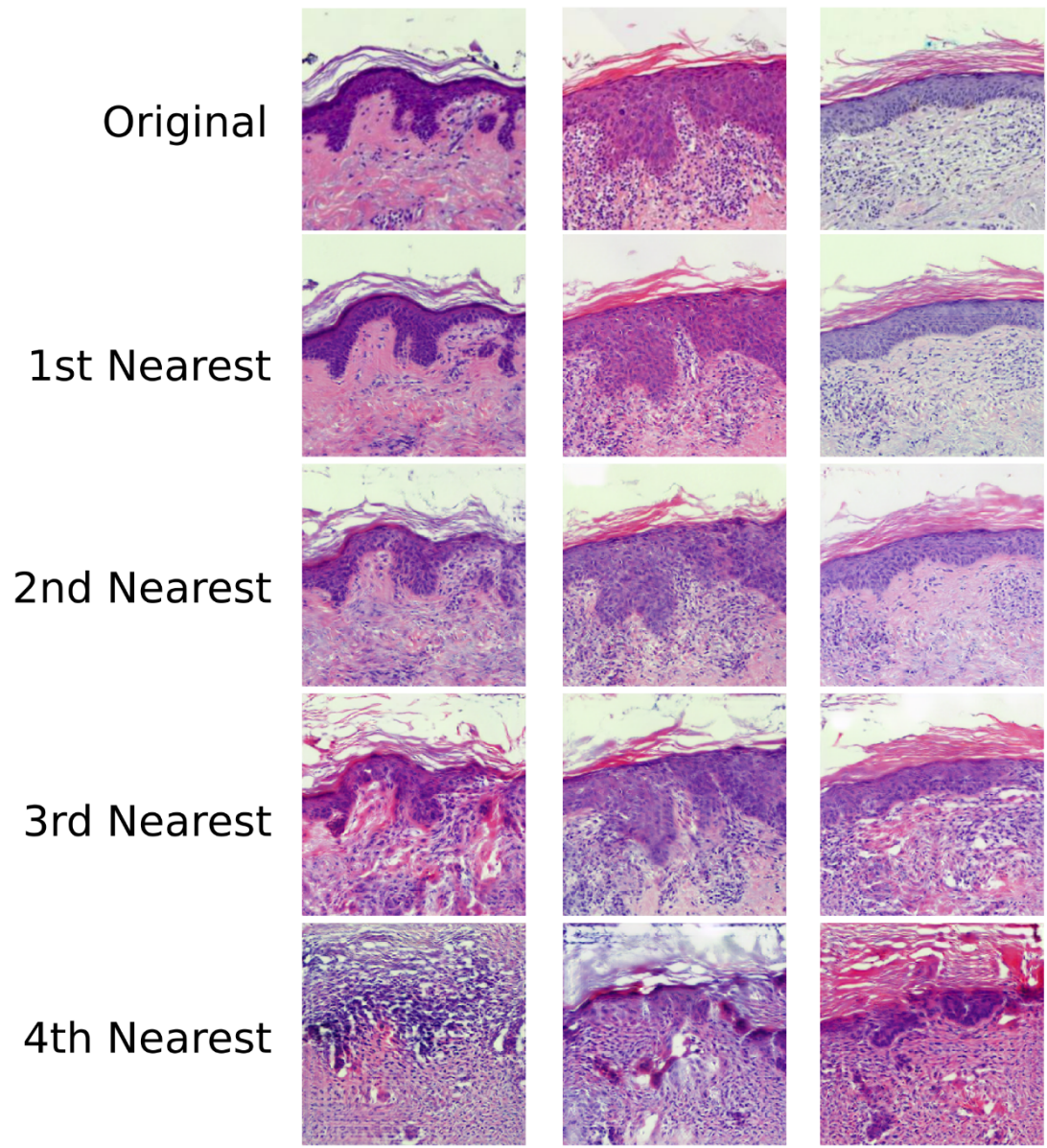

**Fig. S3: Image reconstruction from the k-nearest embedding.** Each image corresponds to a *q* of size 16×16, which leads to an image of size 256×256 pixels. The images are constructed by taking the nearest-k embedding for a given $z_e$ feature. The top 3 images retain large spatial structure with decreasing feature quality. The 4th nearest image shows vague spatial and textural similarities.